\documentclass[10pt,twocolumn,letterpaper]{article}
\pdfoutput=1 
\usepackage{wacv}
\usepackage{times}
\usepackage{epsfig}
\usepackage{graphicx}
\usepackage{amsmath}
\usepackage{amssymb}
\usepackage[accsupp]{axessibility}  
\usepackage{bm}
\newcommand{\wei}[1]{\textcolor{black}{}}

\def\wacvPaperID{435} 

\wacvfinalcopy 

\ifwacvfinal
\def\assignedStartPage{9876} 
\fi

\ifwacvfinal
\usepackage[breaklinks=true,bookmarks=false]{hyperref}
\else
\usepackage[pagebackref=true,breaklinks=true,colorlinks,bookmarks=false]{hyperref}
\fi

\ifwacvfinal
\else
\fi

\begin{document}

\title{Spatial-Temporal Transformer for 3D Point Cloud Sequences}


\author{Yimin Wei\textsuperscript{1,2}, Hao Liu\textsuperscript{1,2}, Tingting Xie\textsuperscript{3}, Qiuhong Ke\textsuperscript{4}, Yulan Guo\textsuperscript{1,2,$*$}\\
\textsuperscript{1}Sun Yat-sen University, \textsuperscript{2}Shenzhen Campus of Sun Yat-sen University, \\ \textsuperscript{3}Queen Mary University of London, \textsuperscript{4}The University of Melbourne\\
{\tt\small weiym9@mail2.sysu.edu.cn,  guoyulan@sysu.edu.cn}
}

\maketitle
\pagestyle{empty}  
\thispagestyle{empty} 

\begin{abstract}
Effective learning of spatial-temporal information within a point cloud sequence is highly important for many down-stream tasks such as 4D semantic segmentation and 3D action recognition. In this paper, we propose a novel framework named \textbf{P}oint \textbf{S}patial-\textbf{T}emporal \textbf{T}ransformer ($PST^2$) to learn spatial-temporal representations from dynamic 3D point cloud sequences. Our $PST^2$ consists of two major modules: a Spatio-Temporal Self-Attention (STSA) module and a Resolution Embedding (RE) module. Our STSA module is introduced to capture the spatial-temporal context information across adjacent frames, while the RE module is proposed to aggregate features across neighbors to enhance the resolution of feature maps. We test the effectiveness our $PST^2$ with two different tasks on point cloud sequences, i.e., 4D semantic segmentation and 3D action recognition. Extensive experiments on three benchmarks show that our $PST^2$ outperforms existing methods on all datasets. The effectiveness of our STSA and RE modules have also been justified with ablation experiments.

\end{abstract}

\section{Introduction}
Point cloud sequences provide both spatial and temporal information of a scene over a period and can be applied in various vision tasks such as 4D semantic segmentation and 3D action recognition. With the rapid development of dynamic point cloud acquisition technologies (such as LiDARs and depth cameras), point cloud sequences are becoming increasingly available and have been investigated for different applications, such as self-driving vehicles, robotics, and augmented reality.


4D semantic segmentation is challenging due to several reasons. First, different from regular and dense 2D image sequences, raw point cloud sequences usually have a regular order in the temporal domain but are unordered in the spatial domain. Besides, point clouds are usually highly sparse. Therefore, it is extremely difficult to model the spatio-temporal structure in a point cloud sequence. Second, it is hard to predict the motion of points within a point cloud sequence. However, the construction of spatio-temporal neighborhoods rely on tracking the motion of points across different frames. It is 
therefore, difficult to effectively extract and aggregate the spatio-temporal context information of points within inter-frame neighborhoods. 

Recently, several methods have been proposed to capture the spatio-temporal dynamics of points from point cloud sequences by utilizing cross-frame information. These methods can be roughly divided into voxel-based methods \cite{p9,p33} and point-based methods \cite{p10,p11,p32,p34}. A voxel-based method converts a 3D point cloud sequence into regular sparse voxels and then extracts pointwise features via a network with 3D convolutions or sparse convolutions. Due to the introduction of quantization error, the performance of this approach is limited. A point-based method constructs a spatial-temporal neighborhood for each seed point (e.g., sampled by Farthest Point Sampling (FPS)), and then applies Multi-Layer Perceptron (MLP) to extract point-wise features. It then passes these features backward and aggregates them with the features of subsequent frames using an inter-frame feature fusion strategy. 
 
\begin{figure*}[ht]
\centering

\includegraphics[width=0.6\linewidth]{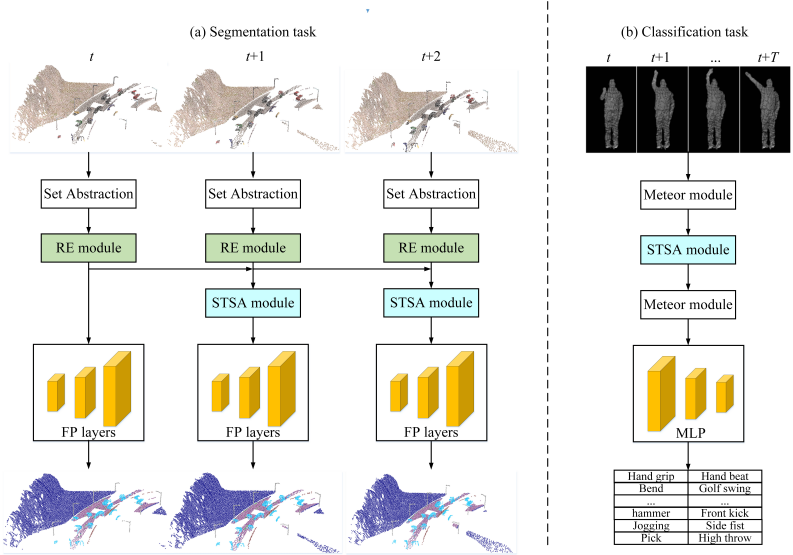}
\caption{{\bf The architecture of our $PST^2$ model.} Our $PST^2$ model is employed for two tasks: 4D semantic segmentation (as shown in (a)) and 3D action recognition  (as shown in (b)). In the 4D semantic segmentation task, we apply two set abstraction layers to extract features, and then employ the Resolution Embedding (RE) module to recover the lost geometric information of the extracted features. Next, we perform convolution on the features of adjacent frames to obtain spatial-temporal patches, and then fed these patches into a Spatio-Temporal Self-Attention (STSA) module to capture spatial-temporal context information. Finally, we use multiple Feature Propagation (FP) layers to produce point-wise semantic predictions. In the 3D action recognition task, we insert our STSA module into the MeteorNet \cite{p10} model to achieve action recognition.}
\label{fig:src2}
\end{figure*}    
    
Although existing point-based methods can avoid the quantization error by directly processing raw point cloud sequences, they still face two major challenges. First, existing methods usually adopt attention operations \cite{p11,p32} or Recurrent Neural Network (RNN) models \cite{p33,p34} to fuse inter-frame features. However, since these methods reply on long-term dependency, the inter-frame information fusion conducted at a frame depends on the fusion of the information of all its previous frames, leading to information redundancy. It is necessary for an inter-frame feature fusion strategy to capture spatio-temporal point context information while reducing the redundancy of previous frames. Besides, current semantic segmentation methods mainly adopt an encoder-decoder architecture. The encoder usually consists of multiple hierarchically stacked feature extraction layers (e.g., the set abstraction layer \cite{p2}) and extracts rich semantic features by reducing the resolution of its feature maps. Therefore, information loss is introduced and the segmentation performance is decreased. 

In this paper, we propose a novel \textbf{P}oint \textbf{S}patial-\textbf{T}emporal \textbf{T}ransformer ($PST^2$) network to tackle the above two challenges. First, we introduce a self-attention based module, i.e., Spatio-Temporal Self-Attention (STSA), to capture inter-frame spatial-temporal context information. More specifically, our $PST^2$ model adaptively aggregates these correlated inter-frame neighborhood features based on the self-attention operation instead of fusing all previous frames. Consequently, the redundancy is reduced, the training speed is increased, and the robustness is improved (with residual connection and layer normalization). Second, we introduce a Resolution Embedding (RE) module to enhance the resolution (which is lost in the encoder-decoder architecture). By aggregating inter-neighborhood features with the attention weights, the resolution can be enhanced and the segmentation performance can be improved significantly.
To further demonstrate the effectiveness of our point spatial-temporal transformer, we introduce a new framework to leverage the proposed STSA module for action recognition from 3D point clouds. Experiments on the Synthia \cite{p17}, SemanticKITTI \cite{p18}, and MSR-Action3D  \cite{p19} datasets demonstrate that our $PST^2$ model outperforms existing methods \cite{p10,p11}. 
    
The contributions of this paper are summarized as follows:
  
  1) We propose a self-attention based module (namely, STSA) to capture point dynamics in point cloud sequences. It is shown that this module can encode the spatio-temporal features from point cloud sequences.
  
  2) We introduce an RE module to enhance the resolution of feature maps. This module clearly improves the segmentation performance.
  
  3) Our $PST^2$ model is tested with two different tasks: 4D semantic segmentation and 3D action recognition. Experiments on three datasets show that the proposed STSA and RE modules can easily be plugged into exiting static point cloud processing pipelines to achieve improved performance.


\section{Related Work}
In this section, we briefly review those works that are highly related to our method, i.e., deep learning method on static point clouds, deep learning method on 3D point cloud sequences, and the transformer.

\subsection{Deep Learning on Static Point Clouds}

 Deep learning has been widely used in many point cloud tasks, such as classification \cite{p1,p2,p52},  part segmentation \cite{p3,p4,p5},  semantic segmentation \cite{p6,p7,p8,p51}, reconstruction \cite{p30,p31}, and object detection \cite{p28,p29,p53}. Existing methods for deep learning on static point clouds can be roughly divided into multi-view based, volumetric-based, and point-based methods. Multi-view based methods \cite{p41,p42} first project a point cloud into multiple views and extract view-wise features, and then fuse these features for static point cloud processing. Volumetric-based methods \cite{p43,p44} convert a point cloud into regular 3D voxels and then employ 3D Convolution Neural Network (CNN) or sparse convolution on the volumetric representation. These two approaches introduce quantization errors and exhibit performance declination, while point-based methods directly work on raw point clouds and reduce the quantization error. 
 
 PointNet \cite{p1} is a pioneering work of point-based methods. Qi et al. \cite{p1} used several MLP layers to extract point-wise local features and adopted a max-pooling layer to aggregate these features. The local features and the aggregated global features are then concatenated to predict the category of each point. In follow-up work, Qi et al. \cite{p2} introduced a hierarchical structure and multi-scale feature learning mechanism into PointNet++, thereby improving the performance of scene semantic segmentation. Subsequent works have been proposed based on the improvement of PointNet++ \cite{p2}. However, these methods cannot directly process dynamic point cloud sequences due to the lack of the inter-frame feature fusion strategy.

\subsection{Deep Learning on 3D Point Cloud Sequences}

In recent years, two major mainstream methods focusing on 3D dynamic point cloud sequences have been explored. 

Several voxelization based methods \wei{\cite{p46,p47,p49}} have been proposed to handle dynamic point cloud sequences. MinkowskiNet et al. \cite{p9} first converted 3D point cloud sequences into 4D occupancy grids, and then employed sparse 4D convolution to process 4D occupancy grids. In 3D Sparse Conv LSTM, Huang et al. \cite{p33} voxelized 3D point cloud sequences into sparse 4D voxels and convolved them on a sparse grid, then passed the memory and hidden features through a sparse convolution inside the LSTM network. The LSTM network produces hidden and memory features that will be passed to the next frame and fused with features of subsequent frames. 

Several methods directly working on raw point cloud sequences have also been explored. MeteorNet \cite{p10} is a leading deep learning method for dynamic 3D point cloud sequences. The structure of MeteorNet \cite{p10} is inherited from PointNet \cite{p1}. By proposing two grouping approaches (including direct grouping and chained-flow grouping) to construct spatio-temporal neighborhoods, inter-neighborhood features are aggregated to capture point dynamics in the point cloud sequence. Cao et al. \cite{p11} proposed an Attention and Structure Aware (ASAP) model, which contains a novel grouping approach (i.e, Spatio-Temporal Correlation) to construct spatio-temporal neighborhoods and a novel attentive temporal embedding layer to fuse the related inter-frame local features in a recurrent fashion. Our \textbf{P}oint \textbf{S}patial-\textbf{T}emporal \textbf{T}ransformer $PST^2$ is also directly performed on raw point cloud sequences, but aims at enhancing the resolution when capturing spatio-temporal dynamics across entire point cloud sequences. \wei{Note that, several methods \cite{p10,p11} and our $PST^2$ employ the same backbone \cite{p1,p2} and focus on the feature fusion strategy. Meanwhile, other works \cite{p48,p50}  focus on introducing new backbones to sample and group those points in local areas more effectively.}

\subsection{Transformer}

Transformer is a typical deep neural network based on the self-attention mechanism and particularly suitable for modeling long-range dependencies. 

Transformer was first proposed in \cite{p12} for the sequence-to-sequence machine translation task, and then extended to various computer vision tasks. In computer vision tasks, image features are obtained from a backbone network and converted into sequences, and then a transformer-based network is employed to process sequence-form features. DEtection TRansformer (DETR) \cite{p13} is a landmark work that considers object detection as a set prediction problem. Specifically, an image is first divided into several candidate regions, and then the object in candidate regions is classified. Recently, transformer has been used in various computer vision applications, including image classification, high/mid-level vision, low-level vision, and video processing. In these applications, Transformer excels to process video sequences and has extended many video tasks, such as video retrieval \cite{p35}, action recognition \cite{p36}, and video object detection \cite{p37}. The Video Instance Segmentation Transformer (VisTR) \cite{p38} extends DETR \cite{p13} for video object instance segmentation (VIS). Specifically, frame-level features are concatenated and fed into a Transformer, and then instance predictions are produced. 

Very recently, Transformer is extended from video tasks to point cloud tasks with promising results.
Guo et al. \cite{p14} proposed a novel transformer-based framework (namely, Point Cloud Transformer (PCT)) for static point cloud learning. They used self-attention to design an offset-attention layer to model irregular point clouds. Zhao et al. \cite{p15} proposed a Point Transformer (PT) layer to fuse point cloud features. 

However, there are few prior Transformer applications \wei{\cite{p45}} to model point cloud sequences. Intuitively, Transformer has the advantage of modeling long-range dependencies and should be an ideal candidate for point cloud sequences tasks. Therefore, we propose the $PST^2$ model to model spatial-temporal information in point cloud sequences with Transformer. 

Our $PST^2$ model is different from the transformers designed in PCT \cite{p15} and PT \cite{p14}. First, our STSA module introduces skip connection and layer normalization operations based on the self-attention operation. Besides, our STSA module divides the input features of the adjacent frame into spatial-temporal patches for further processing, while the input feature is directly processed in \cite{p15,p14}. Consequently, our module is better at capturing the spatial-temporal context information than \cite{p15,p14}. \wei{Moreover, to encode spatial-temporal local features, the Point 4D Transformer (P4Transformer) \cite{p45} was proposed to directly preform 4D convolution in both spatial and temporal domains. 
In contrast, our $PST^2$ model first extracts spatial local features, and then aligns these spatial local features along the temporal domain to generate spatial-temporal patches according to the same seed points. Subsequently, both P4Transformer \cite{p45} and our $PST^2$ model employ self-attention to merge related spatial-temporal local features based on their similarities. }


\section{Point Spatial-Temporal Transformer $PST^2$}
In this section, we first use 4D semantic segmentation as an example to describe our \textbf{P}oint \textbf{S}patial-\textbf{T}emporal \textbf{T}ransformer network $PST^2$, including the overall architecture, the step for spatial-temporal neighborhood construction, the Resolution Embedding (RE)  module, and the Spatio-Temporal Self-Attention (STSA) module. We then adapt the proposed module for 3D action recognition.

\subsection{4D Semantic Segmentation}

\subsubsection{Overview Architecture}
For 4D semantic segmentation, which is a \textit{point-level classification} task, our $PST^2$ model takes a multi-frame point cloud sequence as its input, and predicts the category of each point in the point cloud sequence. Our $PST^2$ model adopts an encoder-decoder architecture. Specifically, the encoder consists of a backbone, an RE module and an STSA module, and the decoder includes multiple Feature Propagation (FP) layers. The FP layer is originally introduced in PointNet++ \cite{p2} to interpolate the semantic features extracted by the encoder upwards and predict the semantic category of each point. 
 
 As illustrated in Fig. \ref{fig:src2} (a), given a point cloud sequence, three steps are adopted to capture point dynamics in our $PST^2$ model. First, we use a backbone network (including two set abstraction layers \cite{p2}) to construct the spatial-temporal neighborhoods and extract point features in each frame (Sec. 3.1.2). Then, we adopt the proposed RE module to enhance the resolution of features in each frame  (Sec. 3.1.3). Finally, we employ the proposed STSA module to fuse the inter-frame features and produce the predictions (Sec. 3.1.3). 

\subsubsection{Spatial-Temporal Neighborhood Construction}

 Given a point cloud sequence $[\bm{S}_1,\bm{S}_2,...,\bm{S}_T]$ of length $T$, each frame of the point cloud is represented as $\bm{S}_t = \{ \bm{p}_i^{(t)}\} _{i = 1}^n$, where $n$ is the number of points. Each point  $\bm{p}_i^{(t)}$ consists of its Euclidean coordinates $\bm{x}_i^{(t)} \in {\mathbb{R}^3}$ and hand-crafted feature $\bm{f}_i^{(t)}$.   A number of seed points $\{ \bm{c}_j^{(t)}\} _{j = 1}^m$ can be sampled from ${\bm{S}_t}$ by the FPS method \cite{p1,p2}, where $m$ is the number of seed points and $m<n$.
 
 We first sample the seed points in the first frame $\bm{S}_1$, and  groups all points within a certain radius around these seed points to form a neighborhood. Next, we use the same seed points in subsequent frames $\{{\bm{S}_2},...,{\bm{S}_T}\}$ to construct spatial-temporal neighborhoods with stable spatial locations at different times. Then, we apply two set abstraction layers  \cite{p2} to extract local features $\bm{h}_i^{(t)} \in {\mathbb{R}^{{s \times d}}}$ from the constructed neighborhood, where $s$ and $d$ are the sizes of the spatial dimension and the feature dimension, respectively.
 

\begin{figure}[ht]
\centering
\includegraphics[width=1.0\linewidth]{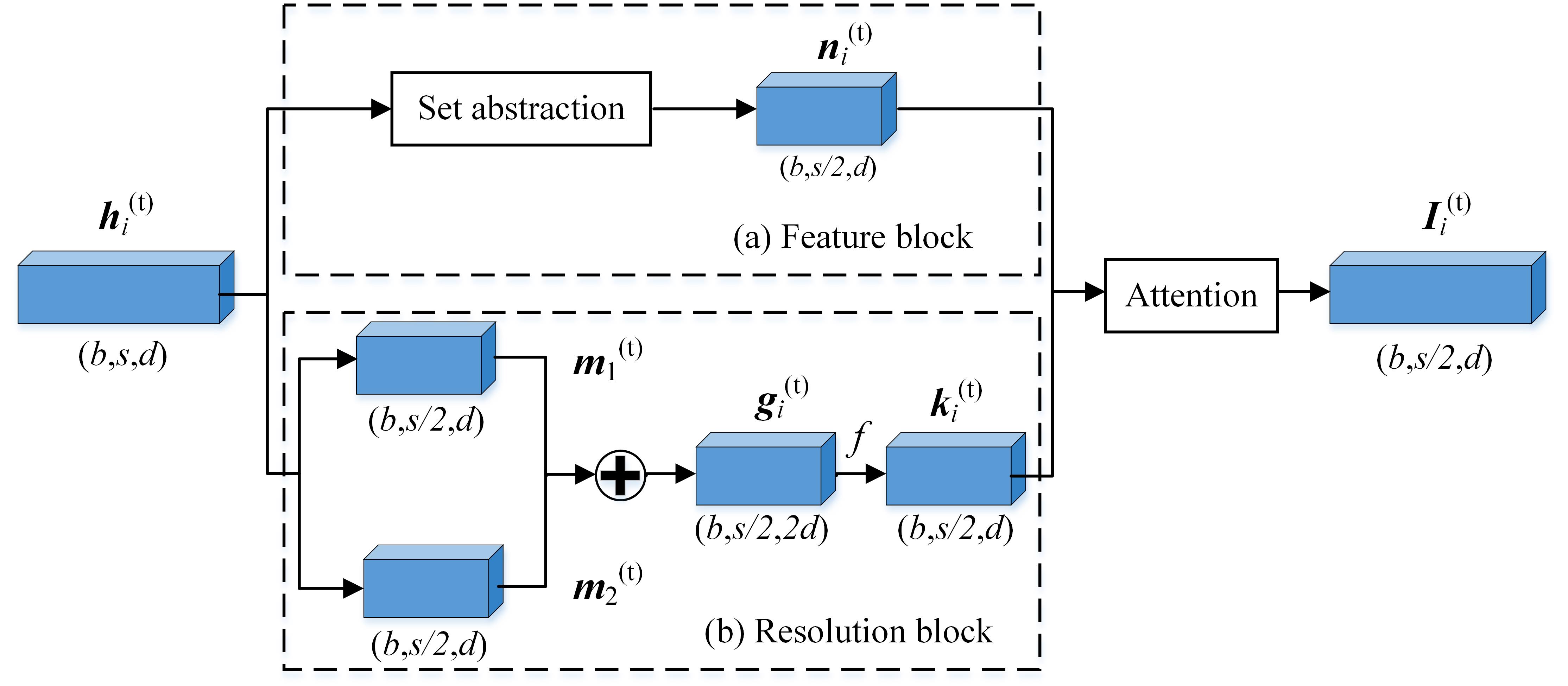}
\caption{{\bf The architecture of our RE module.} Attention and set abstraction stand for the attention operation and the set abstraction layer. 
$b$, $s$ and $d$ represent the batch size, spatial dimension, and feature dimension of the input features, respectively.}
\label{fig:src4}
\end{figure}

\subsubsection{Resolution Embedding (RE)}

 Given the feature representation $\bm{h}_i^{(t)}$, we propose a novel resolution block in the RE module to enhance the resolution of point features in each frame. The RE module consists of a feature block and a resolution block. The feature block was designed to further extract semantic features, while the resolution block extracts the spatial context information from the input features, and treats these context information as an enhanced resolution to incorporate with the features produced by the feature block.

 \textbf{(1) Feature block:} 
  As shown in the Fig. \ref{fig:src4}(a), we perform a set abstraction layer on $\bm{h}_i^{(t)}$ to obtain $\bm{n}_i^{(t)}$. The set abstraction layer is used in the encoder-decoder architecture to hierarchically extract local (even global) features.
    
\textbf{(2) Resolution block:}
As shown in Fig. \ref{fig:src4}(b), we divide $\bm{h}_i^{(t)}$ into $\bm{m}_1^{(t)}$ and $\bm{m}_2^{(t)}$ along the spatial dimension, and then concatenate them along the feature dimension to obtain $\bm{g}_i^{(t)}$. We employ a learnable multi-layer perception $f$ to update the feature dimension of $\bm{g}_i^{(t)}$, resulting in $\bm{k}_i^{(t)}$. This process aims at effectively extracting the spatial inter-neighborhood related information. 
    
 Once $\bm{n}_i^{(t)}$ and $\bm{k}_i^{(t)}$ are obtained, we calculate two scalar attentions $[{a_1},{a_2}]$ according to the correlation of $\bm{n}_i^{(t)}$ and $\bm{k}_i^{(t)}$. By feeding $\bm{n}_i^{(t)}$ and $\bm{k}_i^{(t)}$ into a shared MLP function $\gamma $ and a Softmax function, a weighted-sum feature $\bm{I}_i^{(t)}$ is calculated based on these two attentions: 
\begin{align}
\bm{I}_i^{(t)} = {a_1} \cdot \bm{k}_i^{(t)} + {a_2} \cdot \bm{n}_i^{(t)},
\end{align}
where the attention weights ${a_1}$ and ${a_2}$ are calculated as:
\begin{align}
[{a_1},{a_2}] = Soft\max (\gamma (\bm{k}_i^{(t)},\bm{n}_i^{(t)})).
\end{align}



\begin{figure}[ht]
\centering
\includegraphics[width=1.0\linewidth]{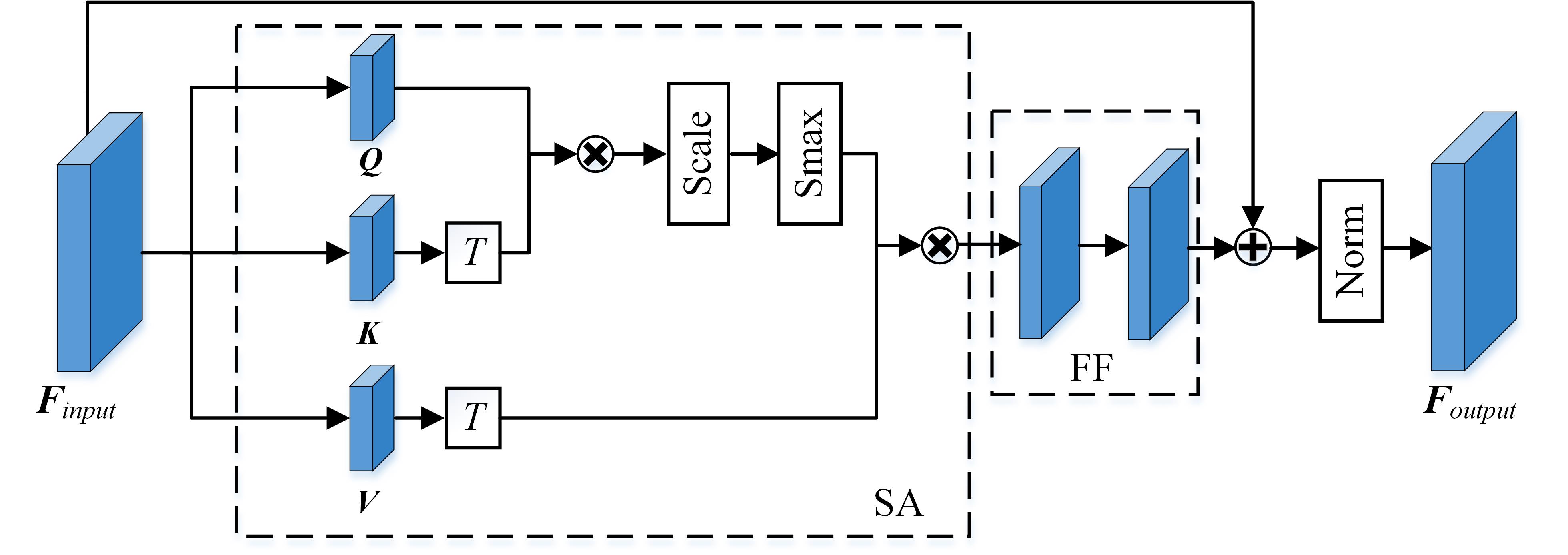}
\caption{{\bf The architecture of our STSA module.} SA, Norm, Smax, $T$ and FF stand for the self-attention operation, the layer normalization function, the softmax operation, the matrix transpose operation, and the feed-forward network, respectively.}
\label{fig:src3}
\end{figure}

\subsubsection{Spatio-Temporal Self-Attention (STSA)}
Given spatial features $\bm{I}_i^{(t)}$ of each frame, the STSA module is used to incorporate inter-frame features along the temporal dimension and capture the spatio-temporal context information. Intuitively, different inter-frame spatial-temporal neighbors contribute differently to the final results. Specifically, for objects with slow-moving speeds, inter-frame spatial-temporal neighbors that are closer in space contributed more than those neighbors that are far apart. In contrast, for objects with fast-moving speeds, inter-frame spatial-temporal neighbors that are far in space should also be considered. Transformer-based methods have the potential to learn the correlation within inter-frame spatial-temporal neighborhoods.
    
As shown in Fig. \ref{fig:src3}, our STSA module consists of two blocks:  spatial-temporal patch division, and  self-attention.

\textbf{(1) Spatial-Temporal Patch Division}    

This block is designed to correlate multiple frame features and generate spatial-temporal patches. Inspired by \cite{p16}, we first divide spatial features $\bm{I}_i^{(t)}$ into patches and then align these inter-frame spatial patches along the temporal domain based on the same seed points to form spatial-temporal patches. These spatial-temporal patches ${\bm{F}_{input}}$ are adjacent along the temporal dimension, and neighborhoods in each patch is consistent along the spatial dimension.


\textbf{(2) Self-Attention} 

The self-attention mechanism measures intra-term semantic affinities within a data sequence by mapping a query and a set of key-value pairs. Following this mechanism, the self-attention block is designed to learn the similarities of the above spatio-temporal patches. Given query matrix $\bm{Q}$, key matrix $\bm{K}$, and value matrix $\bm{V}$, the query, key and value are generated by linear transformations based on the input features ${\bm{F}_{input}} \in {\mathbb{R}^{N \times d}}$:  
   
\begin{align}
(\bm{Q},\bm{K},\bm{V}) = {\bm{F}_{input}} \cdot (\bm{W_q},\bm{W_k},\bm{W_v})
\end{align}
where $\bm{Q},\bm{K},\bm{V} \in {\mathbb{R}^{N \times d}}$, ${\bm{W_q}},\bm{{W_k}},\bm{{W_v}} \in {\mathbb{R}^{{d} \times d}}$ 
are the shared learnable linear transformation implemented by an MLP function, and $d$ is the dimension of the query and key vectors. 

We then use the query and key matrices to calculate the attention weights $\bm{A_1}$: 

\begin{align}
{\bm{A_1}} = {(\bm{a_1})_{i,j}} = \bm{Q} \cdot {\bm{K}^T}
\end{align}

These weights are scaled by $\sqrt{d}$ to produce $\bm{A_2} = {(\bm{a_2})_{i,j}}$:

\begin{align}
\bm{A_2} = {(\bm{a_2})_{i,j}} = \frac{{{{(\bm{a_1})}_{i,j}}}}{{\sqrt {{d_1}} }}
\end{align}

$\bm{A_2}$ is further then normalized to produce $\bm{A_3} = {(\bm{a_3})_{i,j}}$:

\begin{align}
\bm{A_3} = {(\bm{a_3})_{i,j}} = soft\max [{(\bm{a_2})_{i,j}}] = \frac{{\exp [{{(\bm{a_2})}_{i,j}}]}}{{\sum\limits_k {\exp [{{(\bm{a_2})}_{i,k}}]} }}
\end{align}

Then, the features ${\bm{F}_{sa\_out}}$ with self-attention are obtained as the weighted sums of the value vector using their attention weights:
\begin{align}
{\bm{F}_{sa\_out}} = \bm{A_3} \cdot \bm{V}
\end{align}

Finally, the features produced by self-attention ${\bm{F}_{sa\_out}}$  and the input features ${\bm{F}_{input}}$ are further concatenated to obtain the feature ${\bm{F}_{output}}$ through residual connection, a feed-forward network $FeedForward()$, and a layer normalization function $LayerNorm()$:

\begin{align}
{\bm{F}_{output}} = LayerNorm[FeedForward({\bm{F}_{sa\_out}} + {\bm{F}_{input}})]
\end{align}

The self-attention operation introduces randomness in the initial weights and tends to make the output features different from the input feature semantics. To reduce the influence of this randomness and accelerate the model fitting process, skip connection is used to incorporate the input feature and the output feature. Moreover, the difficulty to converge is a generally recognized problem with the original transformer \cite{p12}. Layer normalization is employed to accelerate the model training speed and to improve the robustness of the model by normalizing the output features. 


\subsection{Adaption to 3D Action Recognition}
To further evaluate the ability of $PST^2$ for modeling dynamic point cloud sequences, we also employ $PST^2$ to perform 3D action recognition. 


In a \textit{sequence-level classification} task (e.g., action recognition), our $PST^2$ model takes a multi-frame point cloud sequence as its input, and predicts the label of the sequence (e.g., action category). Since the RE module is specifically designed to enhance the resolution of a segmentation network (implemented with an encoder-decoder structure), we remove the RE module in the classification task. In addition, to intuitively show the effectiveness and easy portability of our STSA module, we improve an existing model MeteorNet \cite{p10} with our STSA module for action recognition. As shown in Fig. \ref{fig:src2}(b), we insert the STSA module into the encoding layer of MeteorNet. The remaining settings are the same as those in \cite{p10}. Note that, the meteor module proposed in \cite{p10} adopts a direct grouping method to roughly group the points in a spatial-temporal neighborhood and further extracts features, which ignore the correlation between these spatial-temporal neighborbors. After inserting our STSA module, more spatial-temporal context information can be obtained and performance can be improved. The effectiveness of the STSA module for a long sequence is evaluated with experimental comparison (see Sec. 4.4). 

\begin{table}[t]
\centering
\newcommand{\tabincell}[2]{\begin{tabular}{@{}#1@{}}#2\end{tabular}}
\caption{\label{tab:table5}{\bf The main hyper-parameters we adopted in the experiments.}}
\begin{tabular}{|c|c|c|c|c|}
\hline

\scriptsize{\bf Experiment} & \scriptsize{\bf learning rate} & \scriptsize{\bf nframe} & \scriptsize{\bf batch size} & \scriptsize{\bf point number}     
\\\hline
\scriptsize{MSR-Action3D} & \scriptsize{0.001} & \tabincell{c}{\scriptsize{4} \\ \scriptsize{8}\\ \scriptsize{12}\\ \scriptsize{16}} &  \tabincell{c}{\scriptsize{16} \\ \scriptsize{8}\\ \scriptsize{8}\\ \scriptsize{8}} &  \tabincell{c}{\scriptsize{2048} \\ \scriptsize{8192}\\ \scriptsize{8192}\\ \scriptsize{10240}}
\\\hline

\scriptsize{Synthia} & \scriptsize{0.0016} & \scriptsize{3} & \scriptsize{2} & \scriptsize{16384}

\\\hline

\scriptsize{SemanticKITTI} & \scriptsize{0.012} & \scriptsize{3} & \scriptsize{2} & \scriptsize{16384}

\\\hline
\end{tabular}
\end{table}

\begin{table*}[t]

\centering
\caption{\label{tab:table1}{{\bf Semantic segmentation results on the Sythia dataset.} Mean accuracy and mean IoU (\%) are used as the evaluation metrics.}}

\resizebox{\textwidth}{!}{
\begin{tabular}{|c|c|c|cc|cccccccccccc|}
\hline

\scriptsize{\bf Method} & \scriptsize{\bf param (M)} & \scriptsize{\bf nframe} & \scriptsize{\bf mAcc} & \scriptsize{\bf mIoU} & \scriptsize{Bldg} & \scriptsize{Road} & \scriptsize{Sdwlk} & \scriptsize{Fence} & \scriptsize{Vegitn} & \scriptsize{Pole} & \scriptsize{Car} & \scriptsize{T.Sign} & \scriptsize{Pdstr} & \scriptsize{Bicyc} & \scriptsize{Lane} & \scriptsize{T.light}

\\\hline
\scriptsize{3D MinkNet \cite{p9}} & \scriptsize{19.31} & \scriptsize{1} & \scriptsize{89.31} & \scriptsize{76.24} & \scriptsize{89.39} & \scriptsize{97.68} & \scriptsize{69.43} & \scriptsize{86.52} & \scriptsize{98.11} & \scriptsize{97.26} & \scriptsize{93.50} & \scriptsize{79.45} & \scriptsize{92.27} & \scriptsize{0.00} & \scriptsize{44.61} & \scriptsize{66.69}
\\

\scriptsize{4D MinkNet \cite{p9}} & \scriptsize{23.72} & \scriptsize{3} & \scriptsize{88.01} & \scriptsize{77.46} & \scriptsize{90.13} & \scriptsize{98.26} & \scriptsize{73.47} & \scriptsize{87.19} & \scriptsize{99.10} & \scriptsize{97.50} & \scriptsize{94.01} & \scriptsize{79.04} & \scriptsize{92.62} & \scriptsize{0.00} & \scriptsize{50.01} & \scriptsize{68.14}
\\\hline

\scriptsize{Pointnet++ \cite{p2}} & \scriptsize{0.88} & \scriptsize{1} & \scriptsize{85.43} & \scriptsize{79.35} & \scriptsize{96.88} & \scriptsize{97.72} & \scriptsize{86.20} & \scriptsize{92.75} & \scriptsize{97.12} & \scriptsize{97.09} & \scriptsize{90.85} & \scriptsize{66.87} & \scriptsize{78.64} & \scriptsize{0.00} & \scriptsize{72.93} & \scriptsize{75.17}

\\\hline
\scriptsize{MeteorNet \cite{p10}} & \scriptsize{1.78} & \scriptsize{3} & \scriptsize{86.78} & \scriptsize{81.80} & \scriptsize{\bf 98.10} & \scriptsize{97.72} & \scriptsize{88.65} & \scriptsize{94.00} & \scriptsize{97.98} & \scriptsize{97.65} & \scriptsize{93.83} & \scriptsize{\bf 84.07} & \scriptsize{80.90} & \scriptsize{0.00} & \scriptsize{71.14} & \scriptsize{77.60}
\\
\scriptsize{ASAP-Net \cite{p11}} & \scriptsize{1.84} & \scriptsize{3} & \scriptsize{87.02} & \scriptsize{\bf 82.73} & \scriptsize{97.67} & \scriptsize{\bf 98.15} & \scriptsize{89.85} & \scriptsize{\bf 95.50} & \scriptsize{97.12} & \scriptsize{97.59} & \scriptsize{94.90} & \scriptsize{80.97} & \scriptsize{\bf 86.08} & \scriptsize{0.00} & \scriptsize{\bf 74.66} & \scriptsize{77.51}

\\
\scriptsize{$PST^2$ (ours)}  & \scriptsize{2.99} & \scriptsize{3} & \scriptsize{\bf 87.03} & \scriptsize{81.86} & \scriptsize{97.48} & \scriptsize{98.12} & \scriptsize{\bf 90.57} & \scriptsize{94.07} & \scriptsize{\bf 98.29} & \scriptsize{\bf 98.07} & \scriptsize{\bf 94.96} & \scriptsize{81.54} & \scriptsize{77.23} & \scriptsize{0.00} & \scriptsize{74.23} & \scriptsize{\bf 77.76}

\\\hline
\end{tabular}}
\end{table*}

\begin{table*}[t]
\centering
\caption{\label{tab:table2}{{\bf Semantic segmentation results on the SemanticKITTI dataset.} Per-class and average IoU (\%) are used as the evaluation metrics.  c1-c19 represent the 19 categories provided in the SemanticKITTI dataset, namely, car (c1), bicycle (c2), motorcycle (c3), truck (c4), other-vehicle (c5), person (c6), bicyclist (c7), motorcyclist (c8), road (c9), parking (c10), sidewalk (c11), other-ground (c12), building (c13), fence (c14), vegetation (c15), trunk (c16), terrain (c17), pole (c18), and traffic-sign (c19).}}
\resizebox{\textwidth}{!}{
\begin{tabular}{|c|c|c|ccccccccccccccccccc|}
\hline

\scriptsize{\bf Method} & \scriptsize{\bf nframe} & \scriptsize{\bf mIoU} & \scriptsize{c1} & \scriptsize{c2} & \scriptsize{c3} & \scriptsize{c4} & \scriptsize{c5} & \scriptsize{c6} & \scriptsize{c7} & \scriptsize{c8} & \scriptsize{c9} & \scriptsize{c10} & \scriptsize{c11} & \scriptsize{c12} & \scriptsize{c13} & \scriptsize{c14} & \scriptsize{c15} & \scriptsize{c16} & \scriptsize{c17} & \scriptsize{c18} & \scriptsize{c19}

\\\hline
\scriptsize{PNv2 \cite{p2}} & \scriptsize{1} & \scriptsize{20.1} & 
\scriptsize{53.7} & \scriptsize{1.9} & \scriptsize{0.2} & \scriptsize{0.9} & \scriptsize{0.2} & \scriptsize{0.9} & \scriptsize{1.0} & \scriptsize{0.0} & \scriptsize{72.0} & \scriptsize{18.7} & \scriptsize{41.8} & \scriptsize{5.6} & \scriptsize{62.3} & \scriptsize{16.9} &\scriptsize{46.5} & \scriptsize{13.8} & \scriptsize{20.0} & \scriptsize{6.0} & \scriptsize{8.9} 

\\\hline
\scriptsize{ASAP-Net \cite{p11}} & \scriptsize{3} & \scriptsize{33.3} & 
\scriptsize{84.1} & \scriptsize{\bf 11.6} & \scriptsize{7.5} & \scriptsize{3.2} & \scriptsize{11.4} & \scriptsize{7.8} & \scriptsize{18.5} & \scriptsize{3.0} & \scriptsize{81.8} & \scriptsize{28.1} & \scriptsize{53.1} & \scriptsize{7.8} & \scriptsize{74.9} & \scriptsize{37.6} &\scriptsize{\bf 64.4} & \scriptsize{27.2} & \scriptsize{51.7} & \scriptsize{22.8} & \scriptsize{\bf 30.8} 

\\\hline
\scriptsize{$PST^2$ (ours)} & \scriptsize{3} & \scriptsize{\bf 36.5} & 
\scriptsize{\bf 84.3} & \scriptsize{3.4} & \scriptsize{\bf 14.7} & \scriptsize{\bf 14.8} & \scriptsize{\bf 14.5} & \scriptsize{\bf 8.7} & \scriptsize{\bf 31.0} & \scriptsize{\bf 22.4} & \scriptsize{81.8} & \scriptsize{\bf 29.6} & \scriptsize{\bf 62.1} & \scriptsize{\bf 14.2} & \scriptsize{\bf 78.8} & \scriptsize{\bf 41.8} & \scriptsize{63.9} & \scriptsize{\bf 30.6} & \scriptsize{\bf 56.6} & \scriptsize{\bf 23.8} & \scriptsize{17.5}

\\\hline

\end{tabular}}
\end{table*}


\begin{table}[t]
\centering
\newcommand{\tabincell}[2]{\begin{tabular}{@{}#1@{}}#2\end{tabular}}
\caption{\label{tab:table3}{\bf 3D action recognition accuracy on the MSRAction3D dataset (\%).}}
\scalebox{1.2}{
\begin{tabular}{|c|c|c|c|}
\hline

\scriptsize{\bf Method} & \scriptsize{\bf Input} & \scriptsize{\bf nframe} & \scriptsize{\bf Accuracy}     
\\\hline
\scriptsize{Vieira et al. \cite{p39}} & \scriptsize{depth} & \scriptsize{20} & \scriptsize{78.20}

\\
\scriptsize{Klaser et al. \cite{p40}} & \scriptsize{depth} & \scriptsize{18} & \scriptsize{81.43}

\\
\scriptsize{Actionlet \cite{p20}} & \scriptsize{skeleton} & \scriptsize{all} & \scriptsize{88.21}

\\\hline
\scriptsize{PointNet++ \cite{p2}} & \scriptsize{point} & \scriptsize{1} & \scriptsize{61.61}

\\\hline
\scriptsize{MeteorNet \cite{p10}} & \scriptsize{point} &
\tabincell{c}{\scriptsize{4} \\ \scriptsize{8}\\ \scriptsize{12}\\ \scriptsize{16}} & 
\tabincell{c}{\scriptsize{78.11} \\ \scriptsize{81.14}\\ \scriptsize{86.53}\\ \scriptsize{88.21}}
\\\hline

\scriptsize{MeteorNet+STSA} & \scriptsize{point} & \tabincell{c}{\scriptsize{4} \\ \scriptsize{8}\\ \scriptsize{12}\\ \scriptsize{16}} &  \tabincell{c}{\scriptsize{81.14 ($\uparrow$ 3.03\%)} \\ \scriptsize{86.53 ($\uparrow$ 5.39\%)}\\ \scriptsize{88.55 ($\uparrow$ 2.02\%)}\\ \scriptsize{\bf 89.22 ($\uparrow$ 1.01\%)}}

\\\hline

\end{tabular}}
\end{table}


\begin{table*}[t]

\centering
\caption{\label{tab:table4}{Ablation results on the SemanticKITTI dataset.  c1-c19 represent the 19 categories provided in the SemanticKITTI dataset, namely, car (c1), bicycle (c2), motorcycle (c3), truck (c4), other-vehicle (c5), person (c6), bicyclist (c7), motorcyclist (c8), road (c9), parking (c10), sidewalk (c11), other-ground (c12), building (c13), fence (c14), vegetation (c15), trunk (c16), terrain (c17), pole (c18), and traffic-sign (c19).}}
\resizebox{\textwidth}{!}{
\begin{tabular}{|c|c|c|c|ccccccccccccccccccc|}
\hline

\scriptsize{\bf Method} & \scriptsize{\bf param (M)} & \scriptsize{\bf nframe} & \scriptsize{\bf mIoU} & \scriptsize{c1} & \scriptsize{c2} & \scriptsize{c3} & \scriptsize{c4} & \scriptsize{c5} & \scriptsize{c6} & \scriptsize{c7} & \scriptsize{c8} & \scriptsize{c9} & \scriptsize{c10} & \scriptsize{c11} & \scriptsize{c12} & \scriptsize{c13} & \scriptsize{c14} & \scriptsize{c15} & \scriptsize{c16} & \scriptsize{c17} & \scriptsize{c18} & \scriptsize{c19}

\\\hline
\scriptsize{PNv2 \cite{p2}} & \scriptsize{3.62} & \scriptsize{1} & \scriptsize{20.1} & 
\scriptsize{53.7} & \scriptsize{1.9} & \scriptsize{0.2} & \scriptsize{0.9} & \scriptsize{0.2} & \scriptsize{0.9} & \scriptsize{1.0} & \scriptsize{0.0} & \scriptsize{72.0} & \scriptsize{18.7} & \scriptsize{41.8} & \scriptsize{5.6} & \scriptsize{62.3} & \scriptsize{16.9} &\scriptsize{46.5} & \scriptsize{13.8} & \scriptsize{20.0} & \scriptsize{6.0} & \scriptsize{8.9} 

\\\hline
\scriptsize{PNv2+RE} & \scriptsize{5.13} & \scriptsize{1} & \scriptsize{\bf 38.6} 
& \scriptsize{\bf 85.4} & \scriptsize{\bf 3.2} & \scriptsize{\bf 13.5} & \scriptsize{\bf 22.7} & \scriptsize{\bf 16.4} & \scriptsize{\bf 9.2} & \scriptsize{\bf 29.1} & \scriptsize{\bf 31.4} & \scriptsize{\bf 82.8} & \scriptsize{\bf 34.1} & \scriptsize{\bf 67.4} & \scriptsize{\bf 13.6} & \scriptsize{\bf 79.4}& \scriptsize{\bf 43.5} & \scriptsize{\bf 64.6} & \scriptsize{\bf 31.2} & \scriptsize{\bf 58.9} & \scriptsize{\bf 24.5}& \scriptsize{\bf 22.6}
 
\\
\scriptsize{PNv2+STSA} & \scriptsize{3.87} & \scriptsize{3} & \scriptsize{\bf 32.4} & 
\scriptsize{\bf 83.4} & \scriptsize{\bf 3.3} & \scriptsize{\bf 7.3} & \scriptsize{\bf 6.4} & \scriptsize{\bf 11.0} & \scriptsize{\bf 6.7} & \scriptsize{\bf 25.9} & \scriptsize{\bf 13.6} & \scriptsize{\bf 78.0} & \scriptsize{\bf 19.1} & \scriptsize{\bf 59.1} & \scriptsize{\bf 9.7} & \scriptsize{\bf 69.1} & \scriptsize{\bf 38.2} &\scriptsize{\bf 59.8} & \scriptsize{\bf 27.0} & \scriptsize{\bf 56.5} & \scriptsize{\bf 23.5} & \scriptsize{\bf 17.4} 

\\\hline

\scriptsize{$PST^2$ (PNv2+STSA+RE)}  & \scriptsize{5.63} & \scriptsize{3} & \scriptsize{\bf 36.5} & 
\scriptsize{\bf 84.3} & \scriptsize{\bf 3.4} & \scriptsize{\bf 14.7} & \scriptsize{\bf 14.8} & \scriptsize{\bf 14.5} & \scriptsize{\bf 8.7} & \scriptsize{\bf 31.0} & \scriptsize{\bf 22.4} & \scriptsize{\bf 81.8} & \scriptsize{\bf 29.6} & \scriptsize{\bf 62.1} & \scriptsize{\bf 14.2} & \scriptsize{\bf 78.8} & \scriptsize{\bf 41.8} & \scriptsize{\bf 63.9} & \scriptsize{\bf 30.6} & \scriptsize{\bf 56.6} & \scriptsize{\bf 23.8} & \scriptsize{\bf 17.5}

\\\hline
\end{tabular}}
\end{table*}
\begin{figure*}[ht]
\centering
\includegraphics[width=0.8\linewidth]{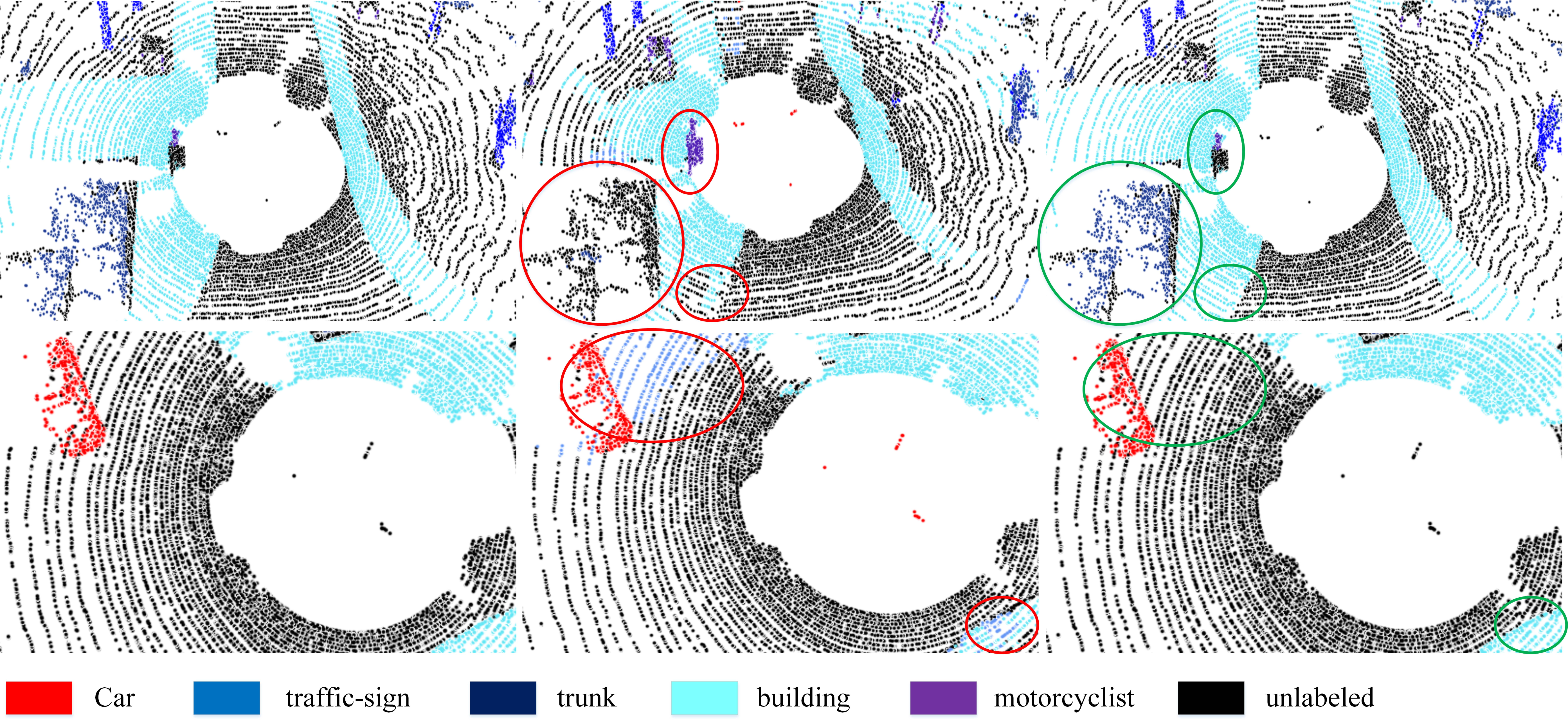}
\caption{{\bf Visualization of 4D semantic segmentation results on the SemanticKITTI dataset.}  From left to right: ground truth, segmentation results of ASAP-Net \cite{p11} and our $PST^2$. Red circles are used to highlight the failure cases produced by ASAP-Net \cite{p11}, while green circles are used to highlight the successful cases produced by our $PST^2$.}
\label{fig:src5}
\end{figure*}


\section{Experiments}

In this section, we first introduce the Synthia \cite{p17}, SemanticKITTI \cite{p18}, and MSRAction3D \cite{p19} datasets and their experimental settings in Sec. 4.1. Then, we report our results on these datasets and compare our model with several existing methods (Secs. 4.2 $\sim$  4.4). Finally, we provide ablation analyses on two spatial-temporal correlation strategies (i.e., STSA and RE) proposed in this paper and justify their contributions in Section 4.5.
\subsection{Datasets and Settings}

We evaluate our $PST^2$ model on the Synthia and SemanticKITTI datasets for 4D semantic segmentation, and the MSRAction3D dataset for 3D action recognition.

\subsubsection{4D Semantic Segmentation}

We conducted two 4D semantic segmentation experiments on the large-scale synthetic dataset Synthia \cite{p17} and the largest publicly available real LiDAR dataset SemanticKITTI \cite{p18}. Then, we compare our $PST^2$ model with the MeteorNet \cite{p10}, ASAP-Net \cite{p11}, and MinkNet \cite{p9} baselines and perform an ablation study on the SemanticKITTI dataset. 

{\bf Synthia} The dataset consists of six sequences of driving scenarios in nine different weather conditions. Each sequence includes 4 stereo RGB-D images from four viewpoints captured from the top of a moving car. Following \cite{p10}, we reconstruct 3D point clouds from RGB and depth images, and then employ the FPS method to downsample the point cloud to 16,384 points per frame. We employ the mean Intersection over Union (mIoU) and the mean Accuracy (mAcc) as the evaluation metrics in our experiments. 

{\bf SemanticKITTI} The dataset provides 23,201 frames for training and 20,351 frames for testing. To compare our method with the state-of-the-art method ASAP-Net \cite{p11}, we follow the multi-frame experiments proposed in \cite{p11}. We only use the mean Intersection over Union (mIoU) as the evaluation metric in our experiments.

\subsubsection{3D Action Recognition}

{\bf MSR-Action3D} The MSR-Action3D dataset consists of 567 depth map sequences acquired by Kinect v1. The depth map sequences include 10 different people, 20 categories of actions, and 23,797 frames. We reconstruct point cloud sequences from these depth maps, and use the same train/test split as previous work \cite{p10,p20}. The overall accuracy is used as the evaluation metric in this experiments.

\subsubsection{Implementation Details}

The main hyper-parameters adopted in the
experiments are reported in Table \ref{tab:table5}. In our $PST^2$ model, we set the number of sampling points in the stacked feature encoding layers as [2048, 512, 128, 64] in the Synthia experiment. Other implementation details, including those for the experiments on Synthia and SemanticKITTI, are set the same as in ASAP-Net \cite{p11}. 
For 3D action recognition, we improve the MeteorNet \cite{p10} baseline with our STSA module, while other settings of the network are the same as MeteorNet \cite{p10}. Moreover, we follow the implementation of FPS provided by PointNet++ \cite{p2}, which is implemented in GPU. All experiments are conducted on an NVIDIA RTX 2080Ti GPU using the ADAM optimizer.

\subsection{Evaluation on Synthia}

Table \ref{tab:table1} reports the semantic segmentation results on the Synthia dataset. Our $PST^2$ consistently outperforms MinkowskiNet \cite{p9}, MeteorNet \cite{p10}, and ASAP-Net \cite{p11}. It establishes a new state-of-the-art result with a mIoU of 81.86 and a mAcc of 87.03. Further, our method achieves the best performance under 5 out of 12 categories. This is mainly caused by the fact that our $PST^2$ model can capture the spatial-temporal context information in a point cloud sequence.

\subsection{Evaluation on SemanticKITTI}

 In the SemanticKITTI experiment, the state-of-the-art method ASAP-Net \cite{p11} adopts PointNet++ (PNv2) \cite{p2} as its backbone network to be incorporated with our ASAP module. For rigorous comparison, we employ the same encoder as ASAP-Net \cite{p11} to incorporate with the proposed STSA and RE module in our $PST^2$.
 
Table \ref{tab:table2} shows the semantic segmentation results on the SemanticKITTI dataset. As shown in Table \ref{tab:table2}, it can seen that our $PST^2$ outperforms PNv2 by a large margin (16.4\% in mIoU). Further, our $PST^2$ surpasses the state-of-the-art method ASAP-Net \cite{p11} by 3.2\% in mIoU and achieves a consistent performance improvement in 15 out of 19 categories. To provide an intuitive comparison, we visualize two segmentation results from the SemanticKITTI dataset, as shown in Fig. \ref{fig:src5}. It can be seen that our $PST^2$ can accurately segment most objects, even in some complex scenes. For example, as can be seen from the first row of Fig. \ref{fig:src5}, most  points on the objects of the motorcyclist category (c8) and the trunk category (c16) are misclassified as unlabeled category by ASAP-Net \cite{p11}, but correctly classified by our $PST^2$.
This is mainly because\wei{,} our RE module can enhance the resolution of the features in each frame, and our STSA module can better incorporate the spatial-temporal context information between  adjacent frames. 


\subsection{Evaluation on MSRAction3D}

Table \ref{tab:table3} presents the classification results on the MSRAction3D dataset. MeteorNet \cite{p10} achieves a significant improvement under the different number of input frames in a sequence by plugging our STSA module. Specifically, given 16 frames in a sequence, 
our classification pipeline achieves the best performance among these methods, with a mAcc of 89.22. Given 8 frames in a sequence, the most remarkable improvement of 5.39\% is achieved by our method. Moreover, it is observed that better classification results can be achieved with more frames in a sequence. That is because, our STSA module can better capture the inter-frame spatial-temporal context information. Given more frames in a sequence, more information can be 
encoded by our method. 

\subsection{Ablation Study}

 In this section, we analyze the impacts of two proposed modules on the semanticKITTI dataset.

\subsubsection{The RE Strategy}

To demonstrate the effectiveness of our RE strategy, we employ two comparative experiments to quantify the contribution of the RE strategy. First, it can be seen from Table \ref{tab:table4} that the PNv2 \cite{p2} network achieves better performances by incorporating our RE-module (i.e., PNv2+RE). That is, the mIoU is improved by ${\rm{18.5\% }}$. Moreover, with the RE strategy, PNv2+RE outperforms PNv2 in all categories, and the performance is improved by up to 38.9\% in the terrain category, and 31.7\% in the car category. Second, 
the RE strategy contributes ${\rm{4.1\% }}$ in terms of mIoU, and achieves better performance in all categories. Specifically, the mIoU gains introduced by  the RE strategy on all these categories vary from 0.1\% to 10.5\%. Large improvements can be observed on parking, building, and motorcyclist. 

\subsubsection{STSA Strategy}
  
  The STSA strategy not only improves the static point cloud pipeline with the ability to process point cloud sequences, but also makes the model achieve better performance. The qualification results are reported in Table \ref{tab:table4}. It can be seen that PNv2+STSA outperforms PNv2 on all categories with our STSA module. Specifically, the performance is improved by ${\rm{12.3\% }}$ in terms of mIoU. The improvement is up to 36.5\% in the terrain category, and 29.7\% in the car category.

\section{Conclusion}

In this paper, we proposed a point spatial-temporal transformer framework to effectively modeling raw 3D point cloud sequences. Our $PST^2$ model consists of two core modules: the spatio-temporal self-attention module and the resolution embedding module. We apply the STSA module to capture inter-frame spatial-temporal context information, and the RE module to aggregate inter-neighborhood features to enhance the resolution. Extensive experiments demonstrate that our $PST^2$ model outperforms existing methods on the tasks of 4D semantic segmentation and 3D action classification. Further, these two modules can be easily plugged into existing static point cloud processing frameworks to achieve remarkable performance improvements.


\section*{Acknowledgement}
This work was partially supported by the National Natural Science Foundation of China (No. U20A20185, 61972435), the Natural Science Foundation of Guangdong Province (2019A1515011271), and the Shenzhen Science and Technology Program (No. RCYX20200714114641140, JCYJ20190807152209394).


{\small
\bibliographystyle{ieee_fullname}
\bibliography{egbib}
}

\end{document}